\documentclass[preprint,12pt]{elsarticle}
\usepackage{amssymb}
\usepackage{amsmath}
\usepackage{makecell} 
\usepackage{flushend}
\usepackage{longtable}
\usepackage{subcaption}
\usepackage{array}        
\usepackage{tabularx}     
\usepackage{xurl}
\usepackage{makecell}     
\newcolumntype{Y}{>{\raggedright\arraybackslash}X} 
\usepackage{float}  

\usepackage{amsmath,amsfonts}
\usepackage{algorithmic}
\usepackage{algorithm}

\usepackage{array}
\usepackage{textcomp}
\usepackage{stfloats}
\usepackage{url}
\usepackage{verbatim}
\usepackage{graphicx}
\usepackage{xcolor}  
 
\usepackage{multirow}  
\usepackage{booktabs} 
\usepackage{lmodern}
\usepackage{multicol}
\usepackage{flushend}

\journal{arVix}

\begin{document}
	
	\begin{frontmatter}
		
		\title{Edge-preserving Image Denoising via Multi-scale Adaptive Statistical Independence Testing} 
		
		\author{Ruyu Yan}
		\author{Da-Qing Zhang\corref{cor1}}
		
		\cortext[cor1]{Corresponding author. Email: d.q.zhang@ustl.edu.cn} 
		
		\affiliation{
			organization={Department of Mathematics, School of Science, University of Science and Technology Liaoning},
			addressline={189 Qianshan Middle Road, Lishan District},
			city={Anshan},
			postcode={114051},
			state={Liaoning Province},
			country={China}
		}
		
		\begin{abstract}
			Edge detection is crucial in image processing, but existing methods often produce overly detailed edge maps, affecting clarity. Fixed-window statistical testing faces issues like scale mismatch and computational redundancy. To address these, we propose a novel Multi-scale Adaptive Independence Testing-based Edge Detection and Denoising (EDD-MAIT), a Multi-scale Adaptive Statistical Testing-based edge detection and denoising method that integrates a channel attention mechanism with independence testing. A gradient-driven adaptive window strategy adjusts window sizes dynamically, improving detail preservation and noise suppression. EDD-MAIT achieves better robustness, accuracy, and efficiency, outperforming traditional and learning-based methods on BSDS500 and BIPED datasets, with improvements in F-score, MSE, PSNR, and reduced runtime. It also shows robustness against Gaussian noise, generating accurate and clean edge maps in noisy environments.
		\end{abstract}
		
		
			
			
			

		
	\end{frontmatter}

	\section{Introduction}
	\label{Introduction}
	Edge detection is a key issue in digital image processing. It not only captures information about object shapes and structures, but also lays the foundation for important applications such as object detection, image segmentation, and object recognition. The goal of edge detection is to accurately mark these areas with large changes in the image. Many existing studies show that edge detection is of great significance in many fields, such as autonomous driving\cite{Barry2019xYOLOAM}, medical imaging\cite{di_jing_medical_2023}, aerospace and image classification\cite{8110709}, facial recognition\cite{Lee17052024,9599570}, and  remote sensing \cite{ma_weixuan_edge_2023}. How to quickly and accurately locate and extract image edge feature information has become one of the research hotspots\cite{zhong2020}. In 2018, Liu et al.\cite{liu2019} proposed a convolutional feature image edge detection extraction algorithm based on five layers of feature diversity; In 2019, Zhao et al.\cite{multifeature2019} proposed a segmentation network for highlighting object detection.

	Although edge detection has long been a core topic in computer vision, traditional methods such as Sobel\cite{2003Pattern}, Prewitt, and Canny\cite{Canny1986A} primarily rely on local gradient operators. These approaches are highly sensitive to noise and often fail in complex scenes, yielding false positives or missing true edges, especially in regions with low contrast or intricate textures. Furthermore, their reliance on fixed-scale processing limits adaptability to images containing multi-scale edge structures, resulting in fragmented or inaccurately localized edges. Most traditional and deep learning-based methods also lack mechanisms to explicitly model contextual information, which hampers their ability to distinguish genuine object boundaries from textures, shadows, or noise.
	
	Recent advances in deep learning have improved edge detection, but challenges remain: reliance on large annotated datasets leads to overfitting, deep network architectures lack interpretability, and high computational costs hinder real-time deployment. To address these issues, we propose the Multi-scale Adaptive Independence Testing-based Edge Detection and Denoising (EDD-MAIT) method. The main contributions are:

	1. Multi-scale Adaptive Sliding Window: The window size is dynamically adjusted based on regional complexity. This approach effectively addresses the varying texture complexities in different regions of actual images, providing a more reasonable balance between detail and global information.
	
	2. Statistical Testing for Noise Detection: By constructing a contingency table with pixel displacement relationships, we apply statistical tests (Fisher or chi-square) to effectively identify noise regions, offering better performance than traditional filtering methods.
	
	3. Dual Thresholding and Complexity-Driven Logic: The Otsu method adaptively sets the thresholds, combined with gradient maps, to preserve high-complexity regions and accurately classify low-complexity areas, minimizing false edge removal.
	
	The rest of this article is structured as follows. The second section reviewed relevant edge detection methods. The third section provides a detailed description of the proposed EDD-MAIT method. The fourth section conducts experimental analysis and quantitative and qualitative evaluations. Finally, the fifth part summarizes this article and outlines future work directions.

	\section{Related Work}
	\label{Related Work}
	\subsection{Traditional edge detection}
	\label{Traditional edge detection}
	Early edge detection methods primarily relied on manual or heuristic-based techniques, which were often time-consuming and illsuited for complex or noisy images. One of the earliest approaches was the Roberts operator, introduced in 1963\cite{1998Edge}, which employed local difference operators to detect intensity discontinuities. However, due to the absence of smoothing and high sensitivity to noise, this method produced coarse and inaccurately localized edges.
	
	Subsequent efforts sought to improve detection accuracy through more robust gradient-based operators. The Sobel\cite{2003Pattern}, Prewitt, and Canny\cite{Canny1986A} operators became foundational in classical edge detection. Among these, the Isotropic Sobel operator\cite{harris1988corner} introduced directionally consistent gradient estimation by incorporating weighted coefficients, enabling improved edge orientation sensitivity. Methods such as Edge Drawing (ED) \cite{Cihan2012Edge} constructed thin, continuous edge segments by tracing anchor points along gradient directions, while EDPF\cite{2012EDPF} employed the Helmholtz principle to achieve fast, parameter-free edge extraction. Liu et al.\cite{Liu2020} further enhanced edge detection robustness by introducing an adaptive thresholding scheme based on the linear correlation between 2D entropy and edge pixel ratios.
	
	Despite their effectiveness, these traditional operators remain inherently sensitive to noise, often resulting in fragmented or discontinuous edge maps\cite{Dhimish2022,Otamendi2023,Sun2020}. To address noise interference, the Laplacian of Gaussian (LoG) operator\cite{2009Fast} incorporated a second-order derivative to detect zero-crossings after Gaussian smoothing. However, its performance was heavily influenced by the scale and shape of the Gaussian kernel. Difference of Gaussians (DoG), proposed by James\cite{Dr2013Difference}, improved noise suppression by emphasizing the contrast between two Gaussian-blurred versions of the image. The Extended DoG (XDoG) method further enhanced edge quality under noisy conditions. Nevertheless, these approaches focused primarily on global edge structures and struggled to capture fine-scale local variations.
	
	Several adaptive techniques have been proposed to improve robustness under varying image conditions. Guo et al.\cite{fpgacanny} demonstrated that adaptive thresholding outperforms fixed-threshold Canny-based approaches under non-uniform illumination. However, such methods often depend on fixed parameters or thresholds, which limits their adaptability and leads to incomplete or broken edge contours.

	Fuzzy logic has also been applied to edge detection due to its capability to model nonlinear and imprecise systems through flexible IF-THEN rule sets\cite{ijsrset}. Unlike knowledge-driven or data-intensive methods, fuzzy systems are easily modifiable and can handle edges of varying sharpness and continuity\cite{miedbsc}. As an extension, Fuzzy Support Vector Machines (FSVM)\cite{rezvani2019intuitionistic} integrate fuzzy logic into SVM frameworks, assigning membership degrees to training samples to enhance noise robustness and reduce false detections, overcoming the sensitivity limitations of traditional SVM-based edge detectors\cite{chen2024islanding}.
	
	Despite these advancements, traditional methods still exhibit limitations in handling complex textures, varying illumination, and noise interference. These challenges highlight the ongoing need for adaptive, data-driven, and context-aware edge detection techniques capable of delivering accurate and robust results across diverse image conditions.
	
	\subsection{Edge detection based on deep learning}
	\label{Edge detection based on deep learning}
	With the rapid development of artificial intelligence, deep learning has emerged as a powerful paradigm for edge detection, addressing many limitations of traditional methods\cite{wei2018fractal}. While classical algorithms have achieved notable progress, they often suffer from limited adaptability to complex scenes and poor inverse performance\cite{wang2020vehicle}. In response, a range of fusion-based edge detection frameworks with enhanced noise robustness have been proposed to improve detection accuracy and stability\cite{shi2017multi}.
	
	Encoder-decoder architectures have been widely adopted in semantic image parsing due to their capacity to model complex spatial hierarchies. The encoder compresses the input into a fixed-length latent representation, while the decoder reconstructs output features, enabling detailed structural inference\cite{polap2018multi}. Recurrent neural networks (RNNs)\cite{sivaraja2020brain} are commonly employed as encoders to capture sequential dependencies. One representative model, the Holistically-Nested Edge Detection (HED) network\cite{yu2021improved}, utilizes deep supervision and multi-scale learning to expand the receptive field, alleviating low-resolution limitations. However, such improvements often come at the cost of increased model complexity and computational demands.
	
	To better mimic human perception, attention mechanisms have been integrated into computer vision tasks since their early application in 1998\cite{1998Edge}. Attention modules have proven highly effective in image classification, object detection, semantic segmentation, and medical imaging\cite{2018Harmonious, Wang2017NonlocalNN, 8954252}. Despite these advantages, edge detection models employing attention mechanisms may still misclassify noise as structural edges in complex backgrounds. Moreover, training such models requires large-scale annotated datasets and substantial computational resources, limiting their practicality.
	
	In addition to neural networks, genetic algorithms (GA) and their variants have also been explored in edge detection. Artificial Neural Networks (ANNs) and deep learning frameworks simulate biological neural systems through highly interconnected computational units, providing flexible modeling of nonlinear systems. Abdel-Gawad et al.\cite{9143122}, for example, applied GA to brain tumor edge detection and achieved an impressive accuracy of 99.09\%, outperforming traditional, fractional-order, and threshold-based methods. Kong et al. \cite{Kong2023} proposed a GA-based adaptive thresholding Sobel operator, effectively enhancing edge continuity and improving upon the limitations of fixed-threshold approaches. A hybrid model based on wavelet transforms and multi-scale processing\cite{xia2022novel} further demonstrated superior noise resilience and edge localization compared to classical operators such as Robert, Sobel, and CNN-based detectors. Additionally, Hu et al. \cite{hu2022distance} introduced a model capable of generating high-quality edge maps and corresponding distance fields, advancing edge-aware representations.
	
	Although these techniques yield significant improvements in detection accuracy, they often face challenges related to high computational complexity and vulnerability to local optima. Deep learning-based methods\cite {2016Richer, 2021RHN} exhibit strong generalization capabilities and precision but demand large volumes of labeled training data, incurring substantial storage and computational overhead. These issues limit their scalability and application in real-time or resource-constrained environments.

	\section{Method}
	\label{The establishment of the IT2FIS-GARCH model}
	This section provides a detailed explanation of the implementation of EDD-MAIT. Initially, we introduce an edge detection approach based on statistical independence testing (including the Chi-square and Fisher tests), which is designed to retain true edge structures while effectively suppressing noise interference. The resulting edge maps exhibit superior precision and robustness, significantly reducing false positives and missed detections that are common in traditional methods under noisy conditions. However, the use of a fixed-size window poses notable limitations, particularly in terms of inadequate adaptability to varying edge scales and redundant computations. To overcome these drawbacks, we propose a multi-scale adaptive window strategy, which dynamically adjusts the local analysis scale according to image complexity, thereby enhancing both detection accuracy and robustness across diverse scenarios.
	
	\subsection{Channel Attention Model}
	\label{Channel Attention Model}
	We extract edge features from the input color image using channel attention. By assigning different weights to each of the three input channels, the model enhances important edge information and improves edge feature extraction. The detailed process of the attention mechanism is illustrated in Fig. \ref{channel attention} and involves the following three steps:
	
	\begin{figure}[]
		\centering
		\includegraphics[width=0.8\textwidth]{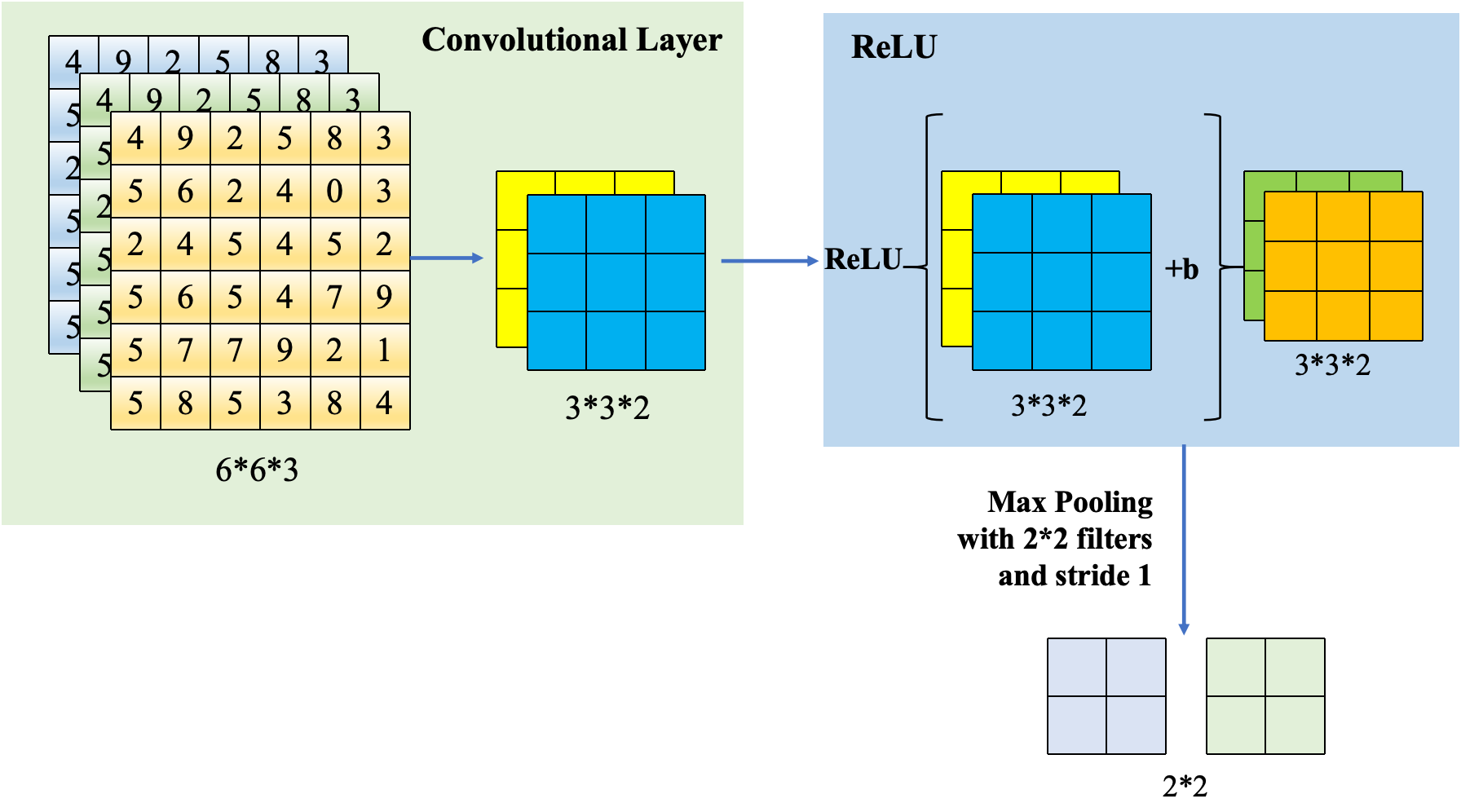}
		\caption{Illustration of Feature Extraction Using Attention Mechanism.}
		\label{channel attention}
	\end{figure}

	The convolution block consists of depthwise and pointwise convolutions. Depthwise convolution extracts local features and is widely used in classic networks such as VGG\cite{10478014}, ResNet\cite{zhao2020deep}, MobileNet\cite{howard2017mobilenets}. Pointwise convolution efficiently fuses channel information while reducing computational cost. This combination balances accuracy and efficiency. In our model, depthwise convolution processes each channel separately, while pointwise convolution integrates features across channels.
	﻿
	To capture complex nonlinear relationships, we apply a nonlinear activation function after the convolution. ReLU is chosen due to its sparse activation, which improves generalization, reduces unnecessary computation, and alleviates the vanishing gradient problem. Its mathematical form is defined as follows:
	
	\begin{equation}
		f(x)=\left\{\begin{matrix}  0&x\le 0 \\  x&x>0\end{matrix}\right.
		\label{relu}
	\end{equation}
	
	The pooling layer is used to reduce the dimension of feature maps and combine weight allocation, preserving the most important information in the image. This operation allows the algorithm to extract key edge information from the input while ignoring non-edge information. In this paper, we use max pooling to filter edge information by outputting the maximum value in each pooling window, obtaining a comprehensive feature representation. The resulting feature map is used in the subsequent boundary determination process based on independence tests. Since the model focuses mainly on edge information in the input, it can still perform well even in the presence of noise or interference, thus enhancing robustness.
	
	\subsection{Preliminary Edge Detection}
	\label{Preliminary Edge Detection}
\begin{figure*}[]
		\centering
		\includegraphics[width=1\textwidth]{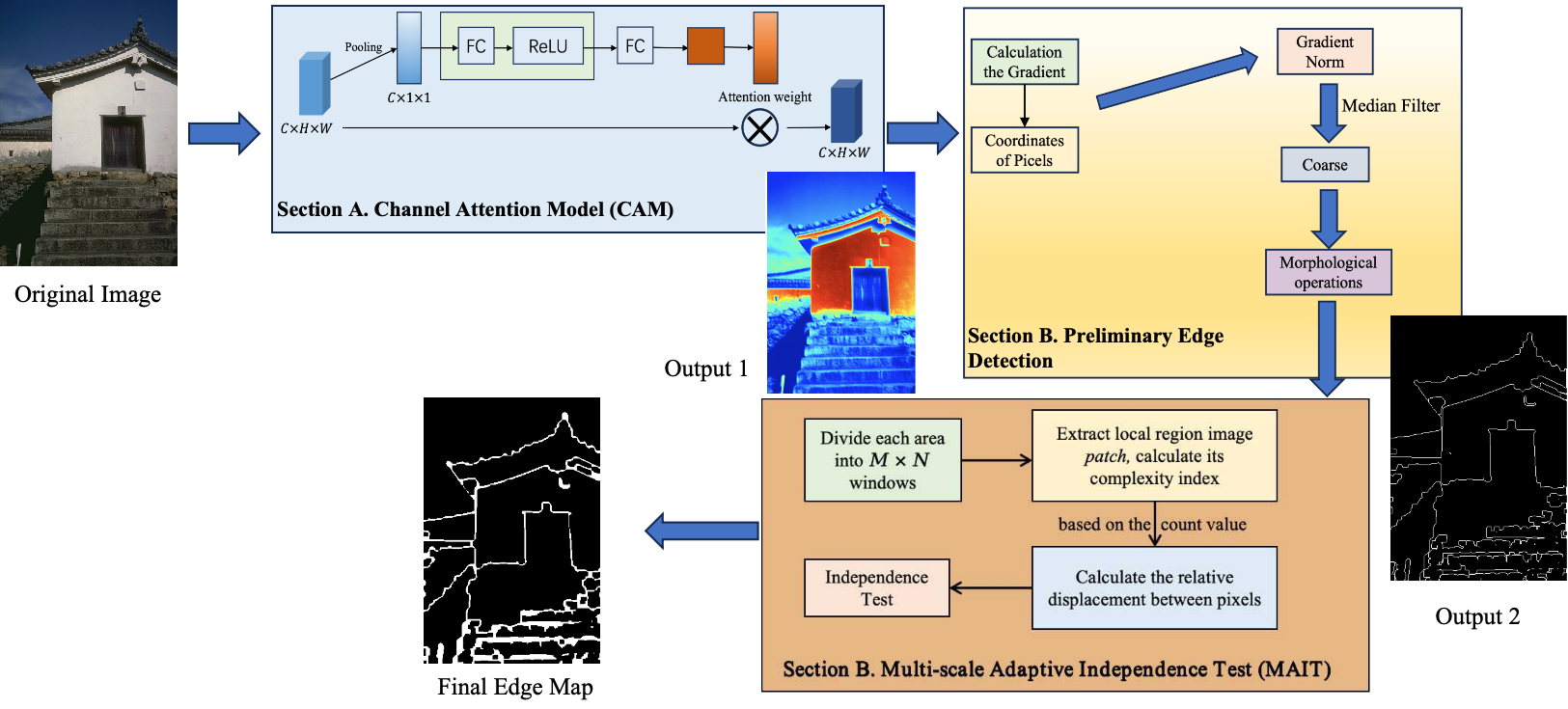}
		\caption{Overview of the proposed EDD-MAIT edge detection framework.}
		\label{framework}
	\end{figure*}

	Through the above operations, we can obtain the initial edge image. To further detect edges and remove noise, we propose a boundary determination method based on independence testing, as shown in Fig. \ref{framework}.
	
	\subsubsection{Calculate gradients and partition membership degrees}
	\label{Calculate gradients and partition membership degrees}
	As discussed in the previous section, we already know that edge detection is the process of detecting significant changes in local pixels of an image. We choose the Sobel operator to calculate the gradient. This operator consists of two sets of $2 \times 2 $ matrices. $G_{x} $ and $G_{y} $ is used in the Sobel operator to detect the edges of the original image $A$ in the horizontal and vertical directions, respectively. Subsequently, these two matrices are convolved with the pixel values of the image to obtain $G_{x} $ and $G_{y} $. The formula is as follows:
	
	\begin{equation}
		G_{x} =\begin{pmatrix}  -1&0  &1 \\  -2& 0 &2 \\  -1& 0 &1\end{pmatrix}\ast  A
		\label{gx}
	\end{equation}
	
	\begin{equation}
		G_{y} =\begin{pmatrix}  1 & 2  &1 \\  0& 0 &0 \\  -1& -2 & -1\end{pmatrix}\ast  A
		\label{gy}
	\end{equation}
	
	Based on the above results, the grayscale value is calculated for each pixel in the image, resulting in a matrix $G$ of the same size as the image $A$: 
	
	\begin{equation}
		G=\sqrt{G_{x}^{2}+G_{y}^{2}  } 
		\label{g}
	\end{equation}
	
	In the context of fuzzy set theory, gradient-based or smooth membership functions serve to represent the degree to which elements belong to fuzzy sets through continuous and differentiable mappings \cite{de_literature_2022}. These functions are well-suited for capturing the gradual transition between full membership and complete non-membership—an essential aspect in edge representation. Among them, the Sigmoid membership function (see Equation \ref{fuzzy1}) exhibits inherent suitability for modeling nonlinear edge transitions. Owing to its monotonic sigmoidal shape, it assigns higher membership values to regions with significant gradient magnitudes \cite{982883, low2020}, while effectively attenuating low-contrast noise. This behavior mirrors the physical process of edge formation, where the strength of intensity gradients corresponds to the perceptual prominence of boundaries.
	
	\begin{equation}
		\mu (x)=\frac{1}{1+e^{-k(x-x_{0} )} } 
		\label{fuzzy1}
	\end{equation}
	where $k$ governs the transition steepness and $x_{0}$ denotes the inflection point. $x_{0}$ typically set as the mean or median of edge gradients. 
	
	This formulation not only preserves edge continuity but also enhances robustness against quantization noise.A smaller $k$ results in a smoother Sigmoid transition, while a larger $k$ makes it steeper. In this paper, we define the pixel gradient value as $x$, the median gradient value as $x_{0}$, and set $k=5$.

	\subsubsection{Median filtering coarsening and morphological smoothing of images}
	\label{Median filtering coarsening and morphological smoothing of images}
	Convolutional filters enhance edge features and suppress noise by exploiting local pixel correlations, assigning weighted values based on intensity differences. For the coarsening process, we adopt the median filter \cite{sambamurthy_scalable_2024}, renowned for its robustness in noise reduction. By computing the median within a local neighborhood, this nonlinear filter effectively eliminates impulsive noise while better retaining edge integrity compared to traditional linear filters. Coarse-graining plays a crucial role in applications demanding high image quality, offering an optimal trade-off between denoising performance and edge preservation.
	
	To further refine image structure, morphological operations such as dilation and erosion \cite{Said_2021} are employed, which serve as essential tools for analyzing and modifying object shapes in binary and grayscale images. In this framework, both coarsening and morphological processing are applied to suppress noise while maintaining structural details. Specifically, $3 \times 3$ structuring elements are used for dilation and erosion, providing a balance between preserving fine local features and maintaining computational efficiency.
	
	\subsection{Adaptive edge detection method based on independence testing}
	\label{Adaptive edge detection method based on independence testing}
	After the above operation, we obtained a smooth edge image, but it still contains some details and noise. Due to the fact that image edges are usually continuous, we have designed an adaptive edge detection method based on independence testing. This method preserves pixels with consistent coordinate changes as the final edge points.
	
	\subsubsection{Calculate the complexity of local regions}
	\label{Calculate the complexity of local regions}
	In traditional image processing methods, a fixed window size is commonly employed across the entire image. However, this approach may lead to information loss in regions with rich details and amplification of noise in relatively smooth regions. To address this limitation, an adaptive window strategy is proposed, which analyzes local gradient magnitudes or other statistical features to dynamically adjust the window size. Specifically, smaller windows are utilized in regions with high local variation to capture fine details, while larger windows are employed in homogeneous regions to suppress noise and reduce false detections.
	
	The core of this strategy lies in the analysis of local gradient amplitude, denoted as $G(x,y)$, which is calculated using the following formulation:
	
	\begin{equation} 
		G(x,y)=\sqrt{\left(\frac{\partial I}{\partial x} \right)^{2}+\left(\frac{\partial I}{\partial y} \right)^{2} } \label{fuzzy} 
	\end{equation}
	
	Based on the computed gradient magnitude $G(x,y)$, the image is classified into two types of regions:
	
	\begin{itemize}
		\item {If $G(x,y) < 0.7$, the region is defined as low or medium complexity. In such cases, statistical independence tests (as detailed in Section 3) are applied to determine whether pixels are noise-contaminated. $W_{\text{min}}$ (set to 8 pixels) is adopted to enhance spatial resolution and prevent edge blurring.}
		
		\item {If $G(x,y) \geq 0.7$, the region is considered high complexity, typically containing edges or textures, and the original pixel values are preserved to maintain detail integrity.}
	\end{itemize}

	To address these challenges, this paper proposes a novel Multi-scale Adaptive Statistical Testing-based Edge Detection and Denoising (EDD-MAIT) method. The key contributions of this work are as follows:
	
	\begin{equation}
		W(x,y)=W_{\text{min}} +(W_{\text{max}} -W_{\text{min}})\cdot \left(1-e^{-\alpha G(x,y)} \right) 
		\label{fuzzy} 
	\end{equation}
	where, $W_{\text{min}}$ and $W_{\text{max}}$ represent the minimum and maximum allowable window sizes, respectively, and the parameter $\alpha$ controls the sensitivity of the window size to changes in gradient magnitude.
	
	This gradient-guided adjustment strategy enables the algorithm to respond adaptively to varying image complexities. In regions with high gradient magnitudes—typically corresponding to edges or textured areas—the exponential term $e^{-\alpha G(x,y)}$ approaches zero, resulting in a window size close to $W_{\text{min}}$. This ensures that fine structural details are preserved and accurately localized. In contrast, for low-gradient areas that represent smooth or homogeneous regions, the exponential term tends toward one, and the window size approaches $W_{\text{max}}$, promoting effective noise suppression and reducing the risk of false edge responses. By continuously adapting the window size in this manner, the proposed method achieves a robust balance between sensitivity to meaningful structures and resilience to noise.
	
	The implementation of this gradient-guided window adaptation strategy can be summarized as follows:
	
	1. High-gradient regions (complex areas such as edges and textures): These regions exhibit rapid changes in pixel intensity and high structural complexity. A smaller window size $W_{\text{min}}$ (set to 8 pixels) is adopted to enhance spatial resolution and prevent edge blurring.
	
	2. Low-gradient regions (smooth areas): These regions display gradual intensity transitions and minimal textural content. A larger window size $W_{\text{max}}$ (set to 64 pixels) is applied to improve the signal-to-noise ratio and suppress false detections caused by noise fluctuations.
	This adaptive mechanism enables the algorithm to maintain high sensitivity and precision in complex regions while ensuring robustness and stability in homogeneous areas.
	
	\subsubsection{Edge Detection Method Based on Independence Test}
	\label{Edge Detection Method Based on Independence Test}
	
	\begin{figure*}[]
		\centering
		\includegraphics[width=0.8\textwidth]{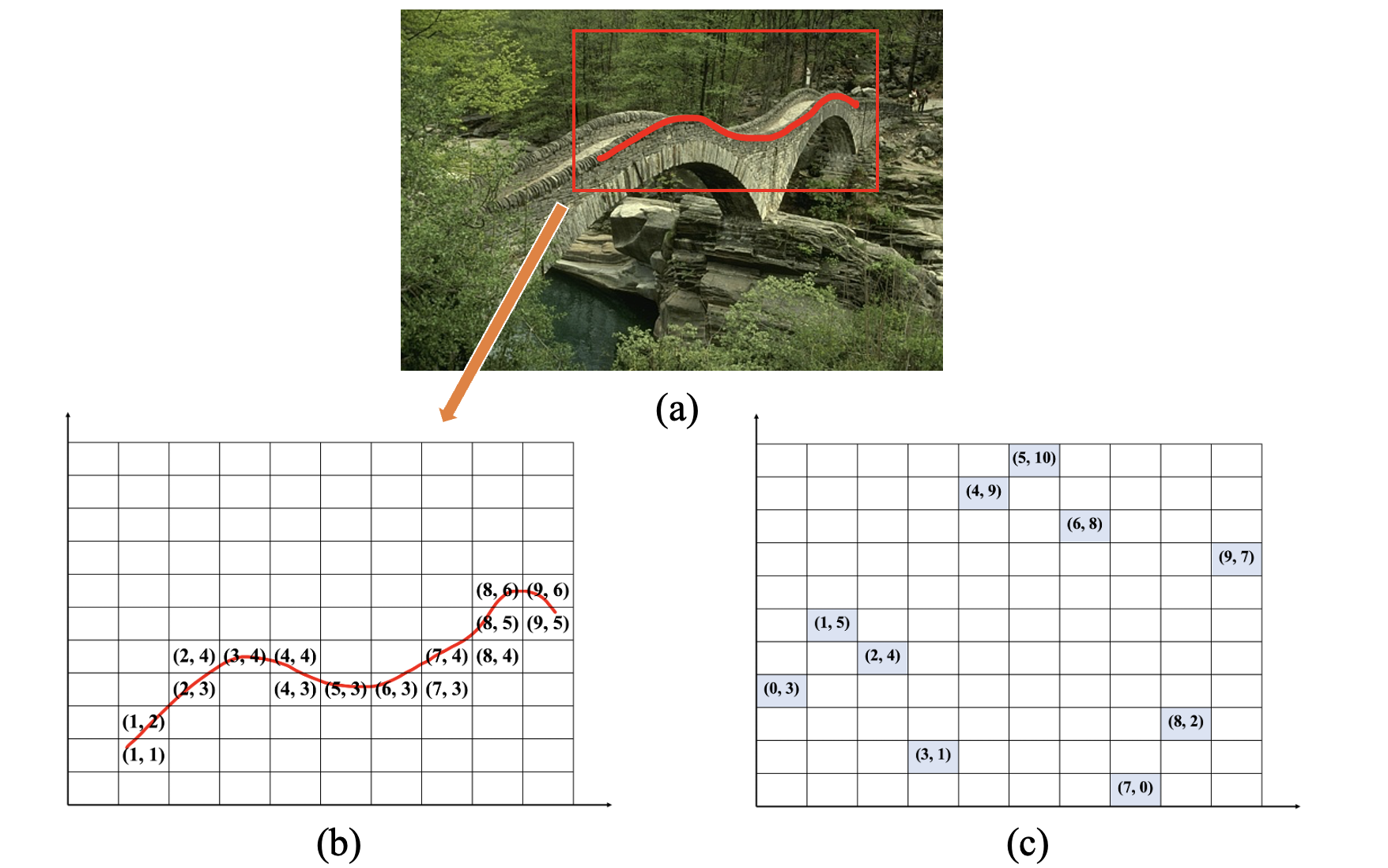}
		\caption{Extract continuous edge curves and randomly generated discrete points that can be regarded as noise points from the image.}
		\label{dlexample}
	\end{figure*}

	Independence testing is a statistical method used to determine whether two categorical variables are independent based on the frequency of events occurring \cite{aaron2017}. In this work, we define the set of edge pixel coordinates in the edge image as $D$. The $x$ and $y$ coordinates of a pixel serve as the categorical variables for the independence test. To analyze the relationship between different pixel coordinates within the set $D$, we compute the relative displacements along the horizontal and vertical axes, denoted as $|\Delta x|$ and $|\Delta y|$, respectively.
	
	\begin{table}
		\begin{center}
			\centering
			\caption{Frequency contingency table of relative displacement among coordinates of different pixels in an edge image}
			\label{bzllb}
			\begin{tabular}{c  c  c  c }
				\hline
				& $\left |  \Delta y \right |\le k$ & $\left |  \Delta y \right |> k$ & total\\
				\hline
				$\left |  \Delta x \right |\le  k$ & $a$ & $b$  & $a+b$\\
				\hline
				$\left |  \Delta x \right |> k$ & $c$ & $d$ & $c+d$ \\
				\hline
				total & $a+c$ & $b+d$ & $n$ \\
				\hline 
			\end{tabular}
		\end{center}
	\end{table}

	For a given pixel $A(x_1, y_1)$ in $D$, we traverse all other points in the set to calculate their relative displacements. For a point $B(x_2, y_2)$ in $D$, the relative displacement in the horizontal direction is $|\Delta x| = |x_1 - x_2|$ and in the vertical direction is $|\Delta y| = |y_1 - y_2|$. To facilitate frequency counting of these displacements, we define a threshold $k$ for the relative displacement. The values of the two variables are categorized into the following ranges: ${|\Delta x| \leq k, |\Delta x| > k}$ for horizontal displacement and ${|\Delta y| \leq k, |\Delta y| > k}$ for vertical displacement. Their corresponding frequency distribution is shown in Table \ref{bzllb}.
	
	In this table, the frequency $a$ represents the count of occurrences where both $|\Delta x| \leq k$ and $|\Delta y| \leq k$. This indicates the number of instances in the set $D$ where the relative displacements in both the horizontal and vertical directions are less than or equal to $k$. The frequencies $b$, $c$, and $d$ are similarly defined for the other displacement categories.

	Let's define the null hypothesis ($H_0$) and the alternative hypothesis ($H_1$) for the independence test of the categorical variables, $x$ and $y$ coordinates:
	
	\begin{itemize}
		\item {$H_0$: The $x$ and $y$ coordinates are independent.}
		\item {$H_1$: The $x$ and $y$ coordinates are not independent, indicating a correlation between them.}
	\end{itemize}

	The frequencies in the contingency table (Table \ref{bzllb}) can fall into two cases: either all frequencies are greater than 5, or at least one frequency is less than 5. Depending on the case, we apply different methods for testing independence: the chi-square test or Fisher's exact test. Both methods are used to assess whether there is independence between two categorical variables.
	
	If all frequencies in the contingency table are greater than 5, we apply the chi-square test. Under the assumption that $H_0$ holds, we calculate the chi-square statistic based on the observed and expected frequencies. The formula for the chi-square statistic is as follows:
	\begin{equation}
		\chi ^{2}=\sum \frac{(O_{ij}-E_{ij}  )^{2} }{E_{ij}}   
		\label{kafang}
	\end{equation}
	where $O_{ij}$ denote the observed frequency, and $E_{ij}$ the expected frequency. This represents the deviation between the observed and expected values. By referring to the $\chi^2$ distribution and the degrees of freedom, the probability of obtaining the calculated statistic under the null hypothesis ($H_0$) can be determined as $p$. If the value of $p$ is smaller than the significance level (typically set at 0.05), we reject the null hypothesis $H_0$ in favor of the alternative hypothesis $H_1$. This implies that the two coordinate variables are not independent, suggesting a consistent pattern in the changes of the coordinates in the set $D$, and therefore, these pixels can be identified as part of the image edges. Conversely, if $p$ is greater than or equal to the significance level, we accept $H_0$, indicating that the two coordinate variables are independent, which implies that the variation between the coordinates is irregular. As such, these points do not contribute to the image edges.

	In the case where any frequency in the contingency table is less than 5, we use Fisher's exact test. Fisher's exact test is based on the hypergeometric distribution, which models the process of drawing $n$ objects from a finite set without replacement, with a specified number of successful outcomes for a particular type of object. Fisher's exact test is particularly suitable for $2 \times 2$ contingency tables, making it appropriate for the binary image processing discussed in this study. The formula for Fisher's exact test is as follows:
	\begin{equation}
		p=\frac{\begin{pmatrix}a+b \\a\end{pmatrix}\begin{pmatrix}c+d \\c\end{pmatrix}}{\begin{pmatrix}n \\a+c\end{pmatrix}} =\frac{\begin{pmatrix}a+b \\b\end{pmatrix}\begin{pmatrix}c+d \\d\end{pmatrix}}{\begin{pmatrix}n \\b+d\end{pmatrix}}
		\label{fisherbz}
	\end{equation}

	This formula calculates the probability $p$ of the observed configuration in the $2 \times 2$ contingency table occurring under all possible scenarios. If $p$ is less than the significance level ($\alpha = 0.05$), we reject the null hypothesis $H_0$ and accept the alternative hypothesis $H_1$, indicating that the coordinate changes of the points in set $D$ follow a regular pattern and can therefore contribute to the image edges. If $p$ exceeds $\alpha$, we accept the null hypothesis $H_0$, concluding that the changes in the point coordinates are random, meaning these points do not define the image edges.
	
	In this method, the red line segment in Fig. \ref{dlexample}(a) marks a continuous curve. Select the area where the red line segment is located in the red box, and then divide the area into $11\times 10 $ regions $\left \{ R_{k}  \right \} _{k=1}^{11\times 10}  $. Figure 8 (b) shows a window with a smooth edge curve and the regions where the pixels are located are marked in the Fig. \ref{dlexample}(b) with coordinates. And Fig. \ref{dlexample}(c) shows a window with noise. The lower left corner is set to (0,0), the horizontal axis is right, and the vertical axis is up.
	
	\begin{table}
		\begin{center}
			\centering
			\caption{Fisher's test contingency table for continuous edge images}
			\label{lianxu1}
			\begin{tabular}{c  c  c  c }
				\hline
				& $\left |  \Delta y \right |\le k$ & $\left |  \Delta y \right |> k$ & total\\
				\hline
				$\left |  \Delta x \right |\le  k$ & 83 & 0  & 83\\
				\hline
				$\left |  \Delta x \right |> k$ & 89 & 10 & 99 \\
				\hline
				total & 172 & 10 & 182 \\
				\hline 
			\end{tabular}
		\end{center}
	\end{table}

	In Fig. \ref{dlexample}(b), all pixel coordinates form the set $D$, where the $x$ and $y$ coordinates of the pixels are treated as two categorical variables. For instance, the coordinates of point $A$ are (2, 3), and for point $B$, they are (1, 1). The relative displacement of point $A$ relative to point $B$ along the horizontal axis is given by $|\Delta x| = |x_1 - x_2| = |2 - 1| = 1$, and the vertical displacement is $|\Delta y| = |y_1 - y_2| = |3 - 1| = 2$. Similarly, the relative displacement between all points can be calculated. We then set a threshold for relative displacement, $k = 3$, and count how often $|\Delta x|$ and $|\Delta y|$ fall within or exceed this limit. After performing this operation, we obtain the contingency Table \ref{lianxu1}. Upon examining the frequency distribution, we observe that when $|\Delta y| > k$ and $|\Delta x| \leq k$, the frequency is less than 5. Consequently, we apply Fisher's exact test to assess the independence of the two categorical variables.

	Using the formula \eqref{fisherbz}, we calculate $p = 0.0021$, which is significantly smaller than the significance level $\alpha = 0.05$. This leads us to reject the null hypothesis $H_0$ and accept the alternative hypothesis $H_1$, indicating that the $x$ and $y$ coordinates of the pixels are not independent. The regularity in the changes of these coordinates suggests that these points form the image edges. Therefore, we classify the pixels in Fig. \ref{dlexample}(a) as edge points.
	
	\begin{table}
		\begin{center}
			\centering
			\caption{Chi-square test contingency table for discontinuous edge images}
			\label{lisan1}
			\begin{tabular}{ c  c  c  c }
				\hline
				& $\left |  \Delta y \right |\le k$ & $\left |  \Delta y \right |> k$ & total\\
				\hline
				$\left |  \Delta x \right |\le  k$ & 21 & 24  & 45\\
				\hline
				$\left |  \Delta x \right |> k$ & 21 & 15  & 36\\
				\hline
				total & 42 & 39 & 81 \\
				\hline 
			\end{tabular}
		\end{center}
	\end{table}

	To assess the method's robustness in the presence of noise, we repeat the analysis on Fig. \ref{dlexample}(c), following the same statistical procedure used in Fig. \ref{dlexample}(b). A contingency table for the categorical variables $\Delta x$ and $\Delta y$ is generated, as shown in Table \ref{lisan1}. We observe that the frequencies in the table are all greater than 5, so we perform a chi-square test. The result of the test yields $p = 0.3722$, which is higher than the significance level. As a result, we accept the null hypothesis $H_0$, suggesting that the $x$ and $y$ coordinates of these points are independent, and the pixel coordinate changes are irregular. Consequently, these points do not contribute to forming a continuous edge. In conclusion, these results confirm the effectiveness of the proposed method in edge detection and denoising.

	\subsubsection{Otsu threshold segmentation}
	\label{Otsu threshold segmentation}
	Otsu thresholding is an adaptive binarization method based on maximizing inter class variance\cite{cheriet1998recursive}, suitable for grayscale images with bimodal distribution. In the context of edge detection, the primary objective of Otsu’s method is to automatically determine a threshold that effectively separates the image into foreground (typically representing objects or features) and background (representing the non-object areas). This is achieved by maximizing the inter-class variance, which quantifies the separation between foreground and background. Assuming the pixel value range of a grayscale image is $\left [ 0,L-1 \right ] $, where $L$ is the grayscale level, setting a certain threshold T to divide the image into background (pixel value $\le$ T) and foreground (pixel value $>$ T), the probabilities of these two parts are represented as:
	
	\begin{equation}
		w_{0} (T)=\sum_{i=0}^{T} p(i)
		\label{kafang7}
	\end{equation}
	
	\begin{equation}
		w_{1} (T)=\sum_{i=T+1}^{L-1} p(i)
		\label{kafang6}
	\end{equation}
	Among them, $p(i)$ is the normalized histogram value at the gray level $i$, which is the proportion of pixels at that gray level in the image. The calculation formulas for background mean and foreground mean are as follows:
	
	\begin{equation}
		\mu _{0}  (T)=\frac{\sum_{i=0}^{T} ip(i)}{w_{0} (T)}
		\label{kafang5}
	\end{equation}
	
	\begin{equation}
		\mu _{1{} } (T)=\frac{\sum_{i=T+1}^{L-1} ip(i)}{w_{1} (T)}
		\label{kafang4}
	\end{equation}
	
	The total mean is defined as:
	\begin{equation}
		\mu _{T} =w_{0} (T)\mu _{0} (T)+w_{1} (T)\mu _{1} (T)
		\label{kafang3}
	\end{equation}
	
	The inter class variance $\sigma ^{2}_{B}(T)$ is given by the following formula:
	\begin{equation}
		\sigma ^{2}_{B}(T) =w_{0} (T)w_{1} (T)(\mu _{0} (T)\mu _{1} (T))^{2} 
		\label{kafang2}
	\end{equation}
	
	The goal of the Otsu method is to find the optimal threshold $T^{\ast } $ that maximizes the inter class variance:
	\begin{equation}
		T^{\ast } =\underset{T}{arg\ max} \ \sigma ^{2}_{B}(T)
		\label{kafang1}
	\end{equation}
	
	In the context of edge detection, the primary objective of Otsu’s method is to automatically determine a threshold that effectively separates the image into foreground (typically representing objects or features) and background (representing the non-object areas). This is achieved by maximizing the inter-class variance, which quantifies the separation between foreground and background.
	
	In the edge detection context, Otsu's method finds the threshold $T^{\ast }$ that maximizes the separation between the foreground and background, identifying strong edge points. A dual-threshold strategy is then used to classify the pixels based on the calculated threshold:
	
	\begin{itemize}
		\item {Pixels with values greater than the high threshold ($T_H$) are directly identified as edge points and are retained.}
		\item {Pixels with values between the low threshold ($T_L$) and the high threshold ($T_H$) are considered as potential edge points that require further verification.}
		\item {Pixels with values below the low threshold ($T_L$) are discarded as they are considered noise.}
	\end{itemize}

	By employing this dual-threshold strategy, the method not only identifies strong edges but also suppresses noise while maintaining potential edge points for further analysis. This enhances the overall edge detection process, ensuring that important features are preserved, and noise is minimized. The algorithm process of this article is shown in Algorithm 1.

	\begin{algorithm}[H]
		\caption{Edge Detection Denoising via Multi-scale Adaptive Statistical Testing}
		\label{alg:EDD-MAIT}
		\begin{algorithmic}
			\STATE \textbf{Input:} Image $A$, filter size $=5$, significance $\alpha = 0.05$, displacement $k=3$, window size $W_{\min}=8$, $W_{\max}=64$, overlap $=0.2$
			\STATE \textbf{Output:} Edge map
			\STATE Extract edge features via attention; compute and normalize gradients
			\STATE Apply sigmoid to obtain membership; use median filtering and morphology to smooth edges
			\STATE Initialize adaptive window and partitioning; generate candidate region set $conditions$
			\FOR{each region in $conditions$}
			\STATE Classify region by gradient magnitude $G(x,y)$
			\STATE Use Otsu thresholding to determine $T_L$, $T_H$
			\IF{$G(x,y) < 0.7$}
			\STATE Use small window $W_{\min}$; 
			\STATE compute contingency table $\{a,b,c,d\}$ via pixel offsets
			
			\FOR{each $(x_1, y_1) \in D$}
			\FOR{each $(x_2, y_2) \in D$}
			\STATE Compute $\left| \Delta x \right|$ and $\left| \Delta y \right|$;
			\IF{$\left| \Delta x \right| \le k$ \AND $\left| \Delta y \right| \le k$}
			\STATE $a \gets a + 1$;
			\ELSIF{$\left| \Delta x \right| \le k$ \AND $\left| \Delta y \right| > k$}
			\STATE $b \gets b + 1$;
			\ELSIF{$\left| \Delta x \right| > k$ \AND $\left| \Delta y \right| \le k$}
			\STATE $c \gets c + 1$;
			\ELSE
			\STATE $d \gets d + 1$;
			\ENDIF
			\ENDFOR
			\ENDFOR
			\STATE Perform region complexity analysis to calculate the region complexity and adjust window size accordingly;
			
			\IF{any of $\{a, b, c, d\} < 5$}
			\STATE Perform Fisher's exact test and filter based on $p$ value;
			\ELSIF{all of $\{a, b, c, d\} > 5$}
			\STATE Perform chi-square test and filter based on $p$ value;
			\ENDIF
			
			\ELSIF{complexity $C > T_H$}
			\STATE Retain original pixels
			\ELSE
			\STATE Use large window $W_{\max}$ and repeat
			\ENDIF
			\ENDFOR
			\STATE \textbf{return} Final edge map
		\end{algorithmic}
	\end{algorithm}

	\section{experiment and result analysis}
	\label{experiment and result analysis}
	\section{Experiment}
	To verify the usability of this method, the algorithm was applied to image edge detection tasks. This section is divided into 5 parts. Firstly, we introduced the public dataset and edge evaluation metrics used in the experiment. Then, the selection of parameters was discussed. And the robustness and computational efficiency of the method proposed in this article were verified. Finally, we evaluated the performance of our method on the BSDS500\cite{bsds500} and BIPED\cite{poma2020dexined} datasets. In addition, we compared F-measure, MSE (mean square error), PSNR (peak signal-to-noise ratio) values, and running time on other algorithms (ED\cite{Cihan2012Edge}, PEL\cite {Akinlar0PEL}, CannySR\cite {7276784}, CannySRPF, EDPF\cite {2012EDPF}, TIP2020\cite {2020An}, MSCNOGP\cite {2024Edge}).

	\subsection{Dataset and edge evaluation metrics}
	
	\subsubsection{Datasets}
	We evaluated the algorithm on two benchmark datasets: BSDS500\cite{bsds500} and BIPED\cite{poma2020dexined}. BSDS500, developed by UC Berkeley, contains 500 images with edge annotations from multiple annotators, and is split into training, validation, and testing sets. It is widely used for general edge detection evaluation. BIPED includes 250 high-resolution natural images (200 for training, 50 for testing), each annotated by five experts and fused to ensure perceptual consistency. Compared to BSDS500, BIPED provides sharper edges and more consistent labels, making it suitable for evaluating fine-grained edge detection in high-resolution scenarios.
	
	\subsubsection{Evaluation metrics}
	We calculated the MSE, PSNR and F-measure value of images processed by different algorithms. MSE represents the similarity between the input and standard images, with a lower MSE indicating better edge detection. The PSNR values further help assess the quality of edge preservation in the processed images. The calculation formula is as follows:
	\begin{equation}
		MSE=\frac{1}{MN}\sum_{i=1}^{M}\sum_{j=1}^{N}[f^{'}(i,j)-f(i,j) ]^{2}  
		\label{mse}
	\end{equation}
	where $f^{'}(i,j)$ and $f(i,j)$ are the detected edge image and the original input image, respectively, while $M$ and $N$ are the width and height of the image.

	PSNR quantifies the degree of image distortion by calculating the ratio of the peak value of the signal to the noise. It is currently another widely used indicator for popularizing image edges. The larger the PSNR value, the smaller the difference between images, and the more accurate the detected image edges. Its calculation formula is as follows: 
	
	\begin{equation}
		PSNR=10log_{10}(\frac{MN\times (\text{max}(L))^{2} }{ {\textstyle \sum_{i=1}^{M} {\textstyle \sum_{j=1}^{N}[f^{'}(i,j )-f(i,j)]^{2} } } } ) 
		\label{psnr}
	\end{equation}

	In Equation~(\ref{psnr}), $max$ is the maximum possible value of the image pixel, and $L$ is the number of bits per sampling point. Generally, the sampling point is 8 bits, i.e. $L=8$. Then max($L$)=255 represents the maximum brightness value of the image.

	Following existing methods\cite{Pu2022Edter}, the F-measure value is frequently used for comparing the performance of edge detection methods. The calculation formula is as follows:
	\begin{equation}
		F=\frac{2\times Presion\times Recall}{Presion+Recall}  
		\label{Fmeasure}
	\end{equation}
	where precision (\(\text{TP} / (\text{TP} + \text{FP})\)) represents the likelihood that an edge pixel predicted by the model is indeed a true edge pixel. Recall (\(\text{TP} / (\text{TP} + \text{FN})\)) measures the proportion of actual edge pixels that are correctly detected. Here, TP, TN, FP, and FN denote the number of true positives, true negatives, false positives, and false negatives, respectively.

	\subsection{Parameter selection}
\begin{figure}[]
		\centering
		\includegraphics[width=0.8\textwidth]{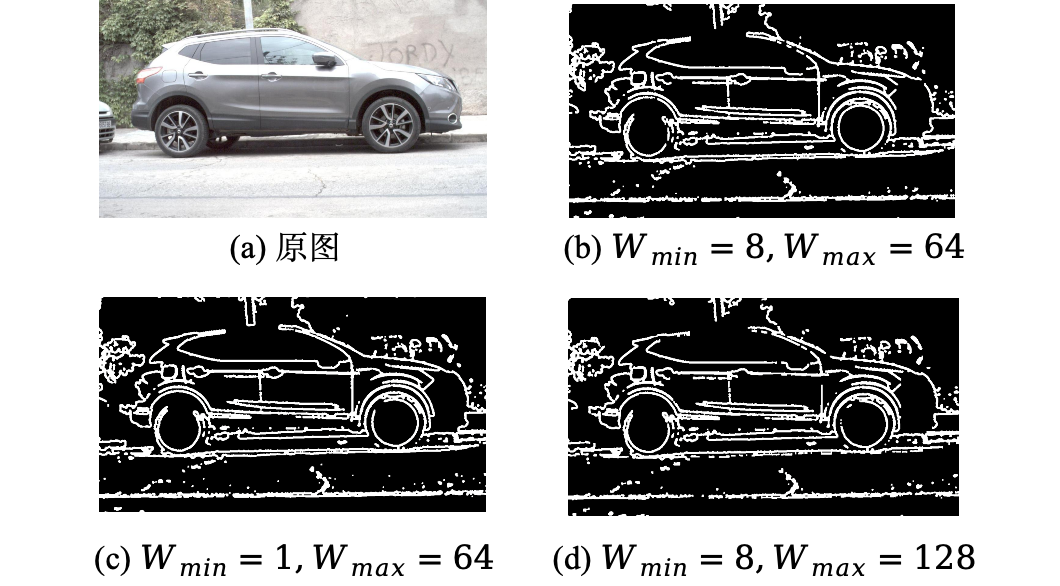}
		\caption{Comparison result diagram of different window sizes}
		\label{chuangkouduibi}
	\end{figure}

	The core parameter design of EDD-MAIT is based on the balanced optimization of local image features and computational efficiency. The specific parameter selection and basis are as follows:
	
	(1) Window size dynamic range: The setting of the minimum $W_{min}=8$ and maximum windows $W_{max}=64$ is derived from the analysis of typical image structures (see section 3). The experiment shows that when the window is smaller than $8\times 8$, the noise interference significantly increases, while when the window is larger than $64\times 64$, the edge localization error will increase, as shown in Fig. \ref{chuangkouduibi}.
	
\begin{figure}[]
		\centering
		\includegraphics[width=3.5in]{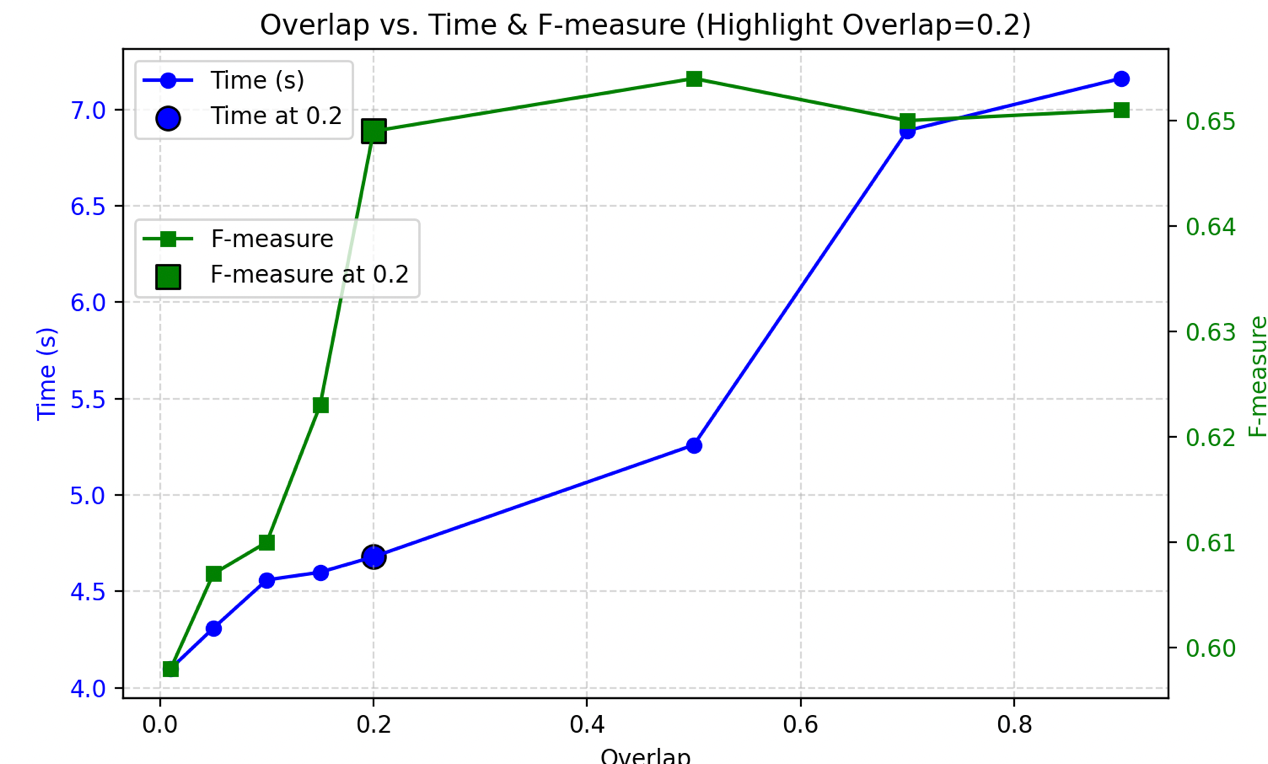}
		\caption{Running time and F-measure corresponding to different overlap rates}
		\label{overlap}
	\end{figure}

	(2) Overlap rate=0.2: This value is determined through experimental balancing of coverage integrity and computational redundancy. When $overlap<0.2$, the F-measure of the BSDS500 dataset decreased by 7.3\%; When $overlap>0.2$, the computation time showed a superlinear increase, as shown in Fig. \ref{overlap}.

	\subsection{Robustness analysis of the EDD-MAIT}
\begin{figure}[]
		\centering
		\includegraphics[width=0.8\textwidth]{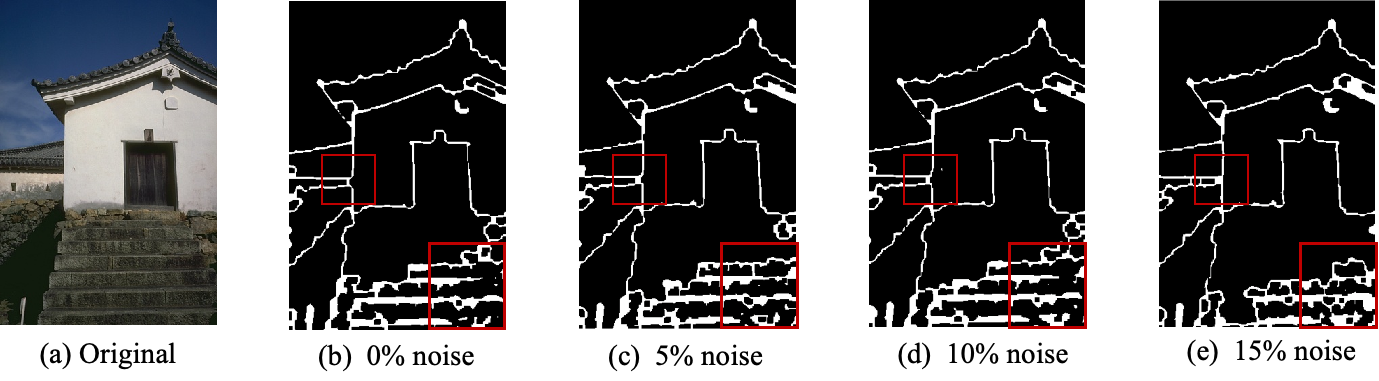}
		\caption{Running time and F-measure corresponding to different overlap rates}
		\label{noise}
	\end{figure}

	To evaluate the robustness of the proposed EDD-MAIT method under noisy conditions, we conducted comparative experiments by introducing Gaussian noise to the input image and observing the corresponding edge detection results. As shown in Fig. \ref{noise}, the first column represents the original clean image, while the second to the fifth columns illustrate the edge maps produced by different methods under Gaussian noise perturbation.
	
	Despite the presence of noise interference, the proposed EDD-MAIT method demonstrates superior robustness and stability. Compared with other methods, EDD-MAIT is capable of preserving true edge structures while effectively suppressing false and fragmented edges caused by noise. This is attributed to two key mechanisms in our method:Channel Attention Mechanism, which enhances edge-relevant features before noise amplifies non-structural information. Multi-scale Adaptive Independence Testing, which dynamically adjusts the local window size, allowing it to distinguish between noise and meaningful structural edges.
	
	In the zoomed-in red boxes, EDD-MAIT maintains clear and continuous contours of architectural edges and stair textures, while other methods suffer from edge blurring, noise amplification, or over-detection.These results confirm that EDD-MAIT is robust against Gaussian noise, and capable of generating accurate and clean edge maps in complex noisy environments.

	\subsection{Comparison of different algorithms}
\begin{figure*}[]
		\centering
		\includegraphics[width=0.8\textwidth]{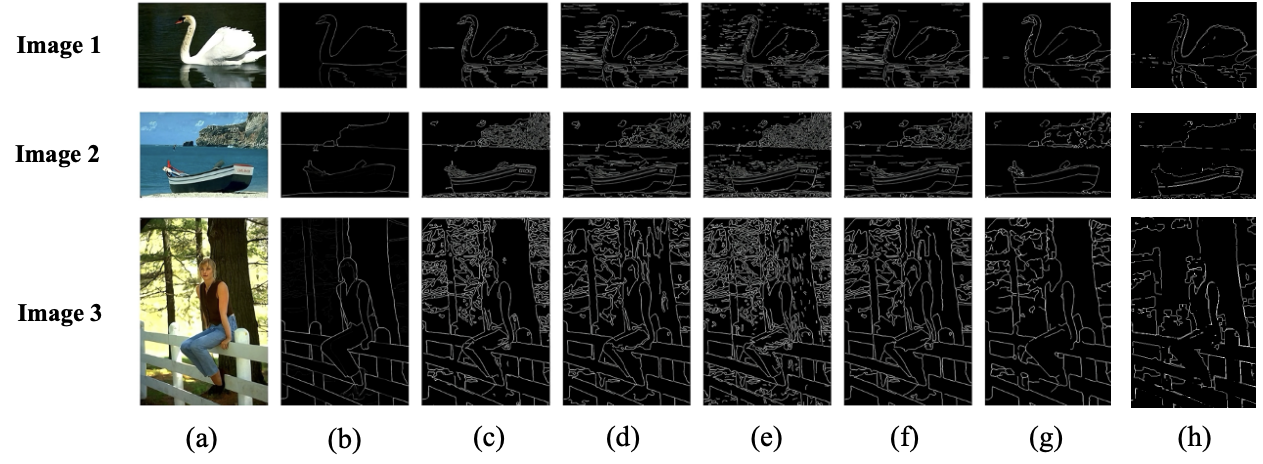}
		\caption{Edge detection comparison image: (a) Original image, (b) Standard edge diagram, (c) CannySR, (d) CannySRPF (e) ED, (f) EDPF, (g) MSCNOGP, (h) EDD-MAIT}
		\label{duibi}
	\end{figure*}

	To verify the superiority of EDD-MAIT, six algorithms were compared on the BSDS500 and BIPED datasets in the experiment. Fig. \ref{duibi} and Table \ref{butongsuanfa} demonstrate the performance of EDD-MAIT and comparative algorithms on the BSDS500 dataset. The key conclusions are as follows:
	
	From the data analysis in Table \ref{butongsuanfa}, it can be seen that the EDD-MAIT proposed in this paper outperforms the recently proposed MSCNOGP in multiple evaluation metrics. Firstly, in terms of mean square error (MSE), EDD-MAIT achieved lower errors in most cases, indicating that it can more effectively reduce the deviation between the reconstructed image and the original image, thereby improving image quality. In Image 1, the MSE of EDD-MAIT is 4.209, which is significantly lower than the 4.653 of MSCNOGP, indicating a reduction in reconstruction error and an improvement in image quality. Similarly, in Images 2 and 3, the MSE of EDD-MAIT further decreased compared to the value of MSCNOGP. In addition, from the perspective of peak signal-to-noise ratio (PSNR), EDD-MAIT achieved 41.799 in Image 1, higher than MSCNOGP's 41.454, further verifying its advantage in image reconstruction quality.
	
	\begin{table*}
		\renewcommand{\arraystretch}{1} 
		\begin{center}
			\centering
			\caption{Evaluation results of images processed by different algorithms}
			\label{butongsuanfa}
			\begin{tabular}{ccccccc}
				\hline
				& image &  Canny &  CannySR &  TIP2020 &  MSCNOGP & EDD-MAIT \\
				\hline
				\multirow{3}{*}{MSE \textdownarrow} & Image 1 & 5.532 & 5.501 & 4.982 & 4.653 & \textbf{4.209} \\
				\cline{2-7}
				& Image 2 & 6.141 & 6.114 & 5.469 & 5.163 & \textbf{5.148} \\
				\cline{2-7}
				& Image 3 & 12.806 & 12.851 & 11.871 & 10.537 & \textbf{10.196} \\
				\hline
				\multirow{3}{*}{PSNR \textuparrow} & Image 1 & 40.702  & 40.726 & 41.362 & 41.502 & \textbf{41.799} \\
				\cline{2-7}
				& Image 2 & 40.284 & 40.267 & 40.360 & 40.952 & \textbf{41.460} \\
				\cline{2-7}
				& Image 3  & 37.057 & 37.042 & 37.345 & 37.913 &  \textbf{38.047} \\
				\hline
			\end{tabular}
		\end{center}
		\footnotesize {The bold represents the best result.}
	\end{table*}

\begin{figure*}[]
		\centering
		\includegraphics[width=0.8\textwidth]{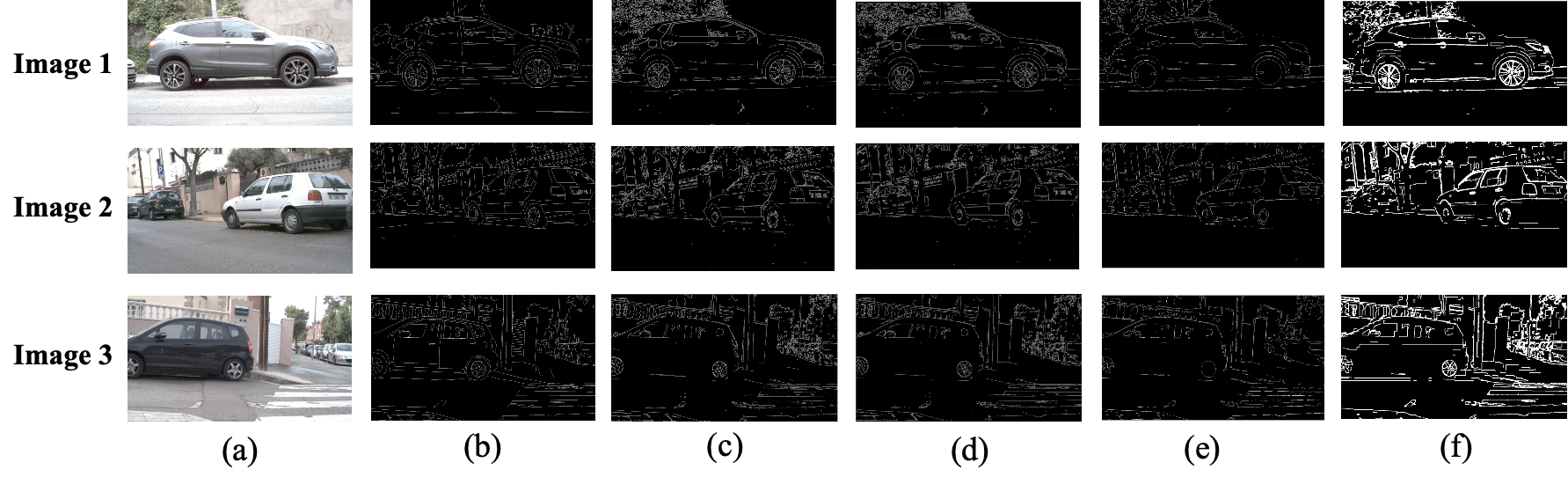}
		\caption{Comparison of edge detection results on the BIPED dataset: (a) Original image, (b) Ground truth, (c) CannySR, (d) MSCNOGP, (f) EDD-MAIT}
		\label{duibi2}
	\end{figure*}

	Fig. \ref{duibi2} and Table \ref{butongsuanfa2} demonstrate the performance of EDD-MAIT and comparative algorithms on the BIPED dataset. From the data in Table \ref{butongsuanfa2}, it can be seen that EDD-MAIT performs well on multiple indicators on the BIPED dataset. In terms of MSE, EDD-MAIT achieved 8.137 on Image 3, which is better than MSCNOGP's 8.764, indicating lower reconstruction error and better image quality. In terms of PSNR, EDD-MAIT is 38.823 on Image 3, higher than MSCNOGP's 38.704, indicating a higher signal-to-noise ratio and better quality of the reconstructed image. Therefore, EDD-MAIT has good performance in MSE and PSNR indicators, especially in providing lower errors and higher signal-to-noise ratio on some images, indicating its better performance in image edge detection.
	
	\begin{table*}
		\renewcommand{\arraystretch}{1} 
		\begin{center}
			\centering
			\caption{Evaluation results of images processed by different algorithms}
			\label{butongsuanfa2}
			\begin{tabular}{ccccccc}
				\hline
				& image &  Canny &  CannySR &  TIP2020 &  MSCNOGP & EDD-MAIT \\
				\hline
				\multirow{3}{*}{MSE \textdownarrow} &  Image 1 & 9.156 & 9.109 & 9.295 & 8.368 & \textbf{7.286} \\
				\cline{2-7}
				& Image 2 & 11.212 & 11.230 & 11.385 & 10.279 & \textbf{8.877} \\
				\cline{2-7}
				& Image 3 & 11.177 & 11.157 & 11.283 & 10.968 & \textbf{8.137} \\
				\hline
				\multirow{3}{*}{PSNR \textuparrow} & Image 1 & 38.514  & 38.536 & 38.449 & 39.218 & \textbf{39.506} \\
				\cline{2-7}
				& Image 2 & 37.634 & 37.627 & 37.567 & 38.024 & \textbf{38.648} \\
				\cline{2-7}
				& Image 3  & 37.648 & 37.655 & 37.606 & 37.931 &  \textbf{38.823} \\
				\hline
			\end{tabular}
		\end{center}
		\footnotesize {The bold represents the best result.}
	\end{table*}

	\subsection{Calculation efficiency analysis}
\begin{figure}[]
		\centering
		\includegraphics[width=0.8\textwidth]{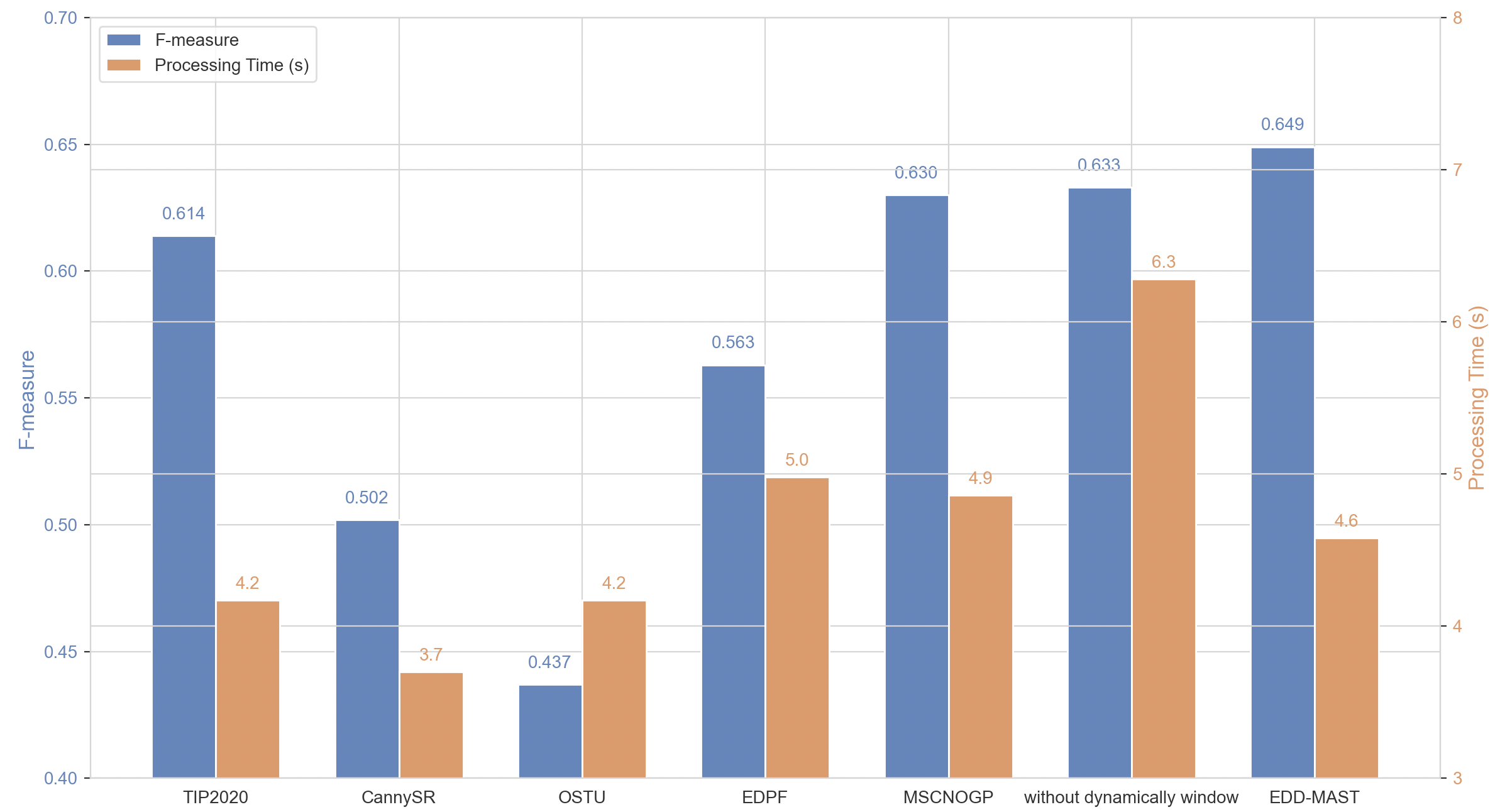}
		\caption{Comparison of running time and F-measure values of different algorithms on the BSDS500 dataset}
		\label{shijian}
	\end{figure}

	The computational efficiency advantage of EDD-MAIT lies in its gradient driven adaptive window strategy, which reduces redundant computation through differentiated resource allocation. The experiment is based on the BSDS500 test set (100 images), and this section analyzes the overall processing speed improvement. As shown in Fig. \ref {shijian}, the EDD-MAIT proposed in this paper outperforms MSCNOGP in terms of efficiency accuracy trade-off: the F-measure value of EDD-MAIT is 0.649, which is higher than MSCNOGP (0.630), indicating a slight improvement in accuracy. In terms of processing time, EDD-MAIT has a processing time of about 4.6 seconds, which is lower than MSCNOGP (4.9 seconds), indicating that the algorithm proposed in this article improves computational efficiency and reduces computational overhead while ensuring accuracy. Therefore, EDD-MAIT significantly reduces computation time while maintaining or even slightly improving segmentation performance, reflecting the advantage of the method in this article in balancing efficiency and accuracy.
	
	In addition, the data in Fig. \ref {shijian} includes the running time without using dynamic adjustment windows, which can be observed to be 6.3 seconds, much higher than the running time of the algorithm in this paper. That is to say, the dynamic adjustment window strategy proposed in this article helps to improve the computational efficiency of the algorithm.
	
	The results of F-measure on the whole testing images of both BSDS500 and BIPED are shown in Fig. \ref{F} and Table \ref{F}. It shows that the EDD-MAIT method achieves 0.649 and 0.463 for BSDS500 and BIPED, respectively. The MSCNOGP and TIP2020 methods followed with second best scores of 0.630 and 0.614 on the BSDS500. This demonstrates our algorithm's high accuracy and robustness in detecting edges of various shapes, sizes, and directions in complex scenes.
	
	﻿
	\begin{figure}[]
		\centering
		\includegraphics[width=0.8\textwidth]{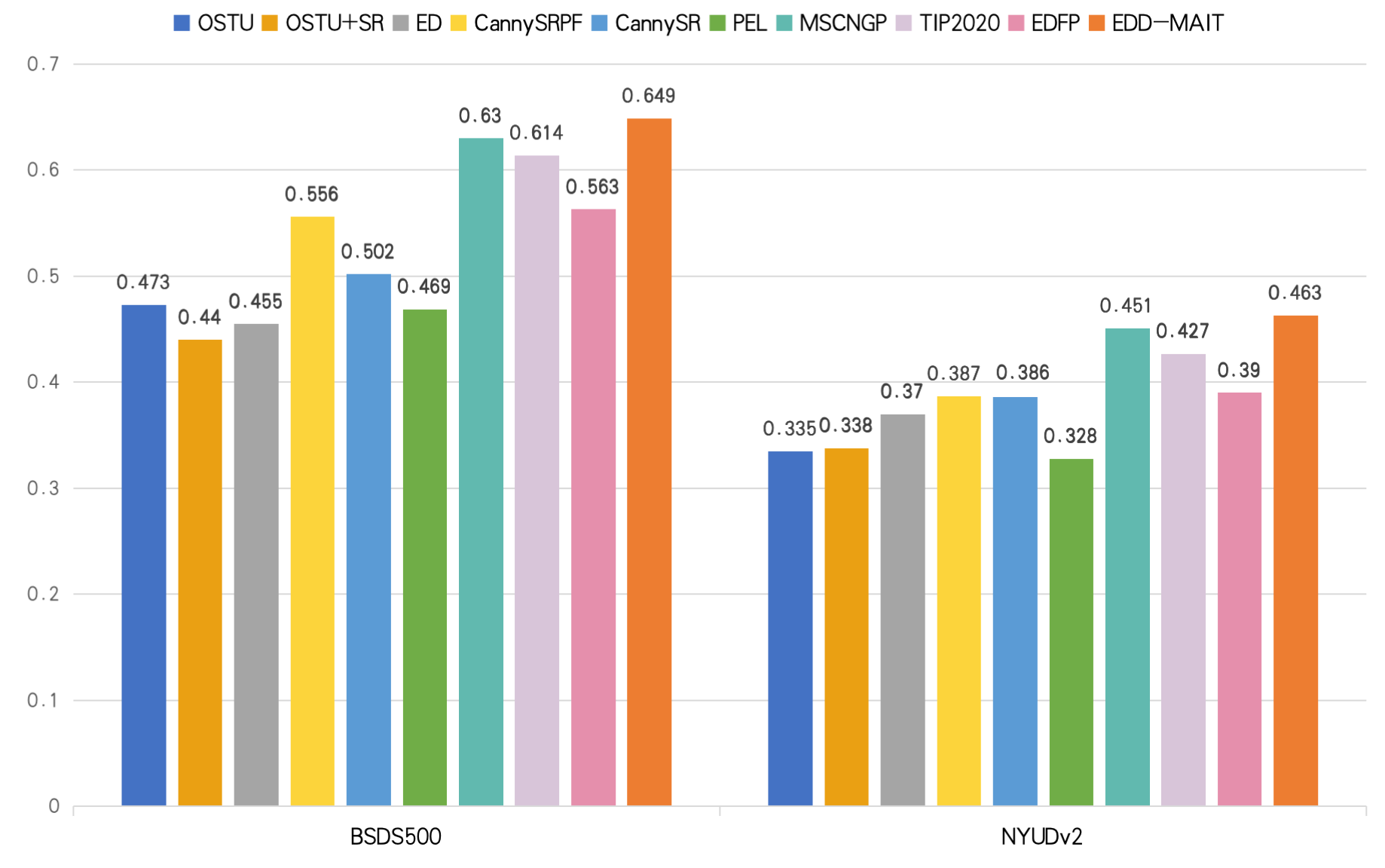}
		\caption{F-measure of the nine methods on both BSDS500 (left half) and BIPED (right half). The nine colors represent the nine methods, respectively.}
		\label{shujuji}
	\end{figure}
	
	\begin{table}
		\begin{center}
			\centering
			\caption{The F-measure of different algorithms on BSDS500 and BIPED datasets}
			\label{F}
			\begin{tabular}{ c  c  c }
				\hline
				& BSDS500 & BIPED \\
				\hline
				OSTU & 0.473  & 0.335\\
				\hline
				OSTU+SR & 0.440 & 0.338 \\
				\hline
				ED \cite{Cihan2012Edge} & 0.455 & 0.370 \\
				\hline
				PEL \cite{Akinlar0PEL} & 0.469 & 0.328 \\
				\hline
				CannySR \cite{Akinlar2015} & 0.502 & 0.386 \\
				\hline
				CannySRPF & 0.556 & 0.387\\
				\hline
				EDPF \cite{2012EDPF} & 0.563 & 0.390\\
				\hline
				TIP2020 \cite{2020An} & 0.614 & 0.427\\
				\hline
				MSCNOGP \cite{2024Edge} & 0.630 & 0.451\\
				\hline
				EDD-MAIT & \textbf{0.649} & \textbf{0.463}\\
				\hline 
			\end{tabular}
		\end{center}
	\end{table}

	\section{Conclusion}
	\label{Conclusion}
	In this paper, we proposed a novel edge detection denoising algorithm based on multi-scale adaptive Independence testing(EDD-MAIT). By integrating channel attention mechanisms with gradient-based feature extraction, the method effectively enhances edge representation. The adoption of a sigmoid-based membership function allows for partitioning of edge regions, enabling more flexible and accurate identification of potential edge pixels. Furthermore, the algorithm employs a complexity-driven window adjustment strategy, dynamically selecting between small and large analysis windows. By incorporating both Fisher's exact test and the chi-square test according to the observed data, the method robustly distinguishes true edges from noise, even in complex or low-gradient regions. The dual-threshold approach ensures adaptive processing across regions of varying texture and detail levels.
	
	Despite the promising overall performance of EDD-MAIT, there remains room for further improvement. For instance, combining spatial attention with channel attention may enhance the method's responsiveness to weak edges in complex backgrounds. In addition, to address the slight discontinuity of edges in extremely low signal-to-noise ratio images, incorporating structural priors or graph-based optimization techniques could be considered. Future research may also explore the application of EDD-MAIT in specific domains such as video sequences and medical imaging to further extend its practicality and generalization capabilities.
	
	\appendix
	
	\bibliographystyle{elsarticle-num}  
	\bibliography{reference} 
\end{document}